\documentclass[preprint,3p,times,11pt]{elsarticle}

\usepackage{graphicx,lineno}
\usepackage{amsmath,amssymb,mathtools,underscore,subfigure}
\usepackage{bbold}
\usepackage[colorlinks]{hyperref}
\usepackage[utf8]{inputenc}
\usepackage{varioref}
\usepackage{url,longtable,afterpage}
\AtBeginDocument{}

\DeclareFontFamily{U}{rcjhbltx}{}
\DeclareFontShape{U}{rcjhbltx}{m}{n}{<->rcjhbltx}{}
\DeclareSymbolFont{hebrewletters}{U}{rcjhbltx}{m}{n}
\DeclareMathSymbol{\shin}{\mathord}{hebrewletters}{152}

\newcommand{\abs}[1]{{{\left | {#1} \right |}}}
\newcommand{\skipall}[1]{}
\newcommand{\my}[0]{\ensuremath{\mu}}

\newcommand{\mline}[0]{\middle\vert}

\DeclareMathOperator{\cross}{\times}

\DeclareMathOperator{\sg}{sg}

\newcommand{\eqdef}{\coloneqq}
\renewcommand{\phi}{\varphi}
\renewcommand{\epsilon}{\varepsilon}
\begin{document}

\begin{frontmatter}
\journal{just arXiv so far}
\title{Batchless Normalization: How to Normalize Activations Across Instances with Minimal Memory Requirements}

\newcommand{\authors}[1]{#1}

\authors{
\author[label1]{Benjamin Berger}
\address[label1]{Leibniz Universität Hannover}
\author[label2]{Victor Uc Cetina}
\address[label2]{Universidad Autónoma de Yucatán}
}

\begin{abstract}
In training neural networks, batch normalization has many benefits, not all of them entirely understood. But it also has some drawbacks.
Foremost is arguably memory consumption, as computing the batch statistics requires all instances within the batch to be processed simultaneously, whereas without batch normalization it would be possible to process them one by one while accumulating the weight gradients.
Another drawback is that that distribution parameters (mean and standard deviation) are unlike all other model parameters in that they are not trained using gradient descent but require special treatment, complicating implementation. In this paper, I show a simple and straightforward way to address these issues. The idea, in short, is to add terms to the loss that, for each activation, cause the minimization of the negative log likelihood of a Gaussian distribution that is used to normalize the activation. Among other benefits, this will hopefully contribute to the democratization of AI research by means of lowering the hardware requirements for training larger models.
\end{abstract}
\end{frontmatter}
%\linenumbers
%\allowdisplaybreaks
\section{Introduction}
Batch normalization\footnote{Batch normalization actually performs standardiazation and not normalization, but the name stuck.}\cite{ioffe2015batch} is a popular technique for improving the training of neural networks. The basic idea is to take a look at each activation after a layer and to normalize it by scaling and shifting it so that the mean and standard deviation across the current batch for that activation become $0$ and $1$, respectively. This is supposed to approximate a normalization with the population statistics by means of the batch statistics, leading to approximately normalized inputs for the following layer. That being said, a batch normalization layer is usually assumed to include a denormalization afterwards, that is, the normalized activations are once again transformed affinely so as to have a certain mean and standard deviation, which are learnable parameters of the model. This means that the inputs to the next layer are not normalized, but rather conform approximately to a mean and standard deviation that are independent of whatever the layer before the batch normalization layer produced.

The benefits of batch normalization are manifest empirically, but their theoretical understanding is under debate. I will say no more about this as my intention is not to criticize the benefits, but to address the shortcomings of which there are also several:
\label{bnsc}
\begin{description}
\item[Memory consumption:] All instances of the batch must be in memory at the same time in order to compute the batch statistics. This can become a problem if the data required per instance (the activations as well as the gradients of the activations with respect to loss) do not fit on the available hardware multiple times. Even if multiple devices are available, it requires either communication between these at each batch normalization layer, or to compromise on the accuracy of the batch statistics by computing it separately and independently for each device.
\item[Implementation issues:] The statistical parameters for a batch normalization layer are not learned using gradient descent and some optimizer, like everything else. After the network has been trained, there should be extra passes over all training data to determine the statistics parameters for the entire training data. Since that can be quite expensive, moving averages of the batch statistics parameters are typically computed during training and used during inference.
\item[Train-test-discrepancy:] This in turn means there is a difference between the data a network layer sees during training and the data it sees during testing. The activations in the network after training follow a slightly different distribution from how they were when it was trained.
\item[Cheating:] The instances in the batch are no longer processed independently from each other if batch normalization is present. This can lead to cheating in some settings such as contrastive learning\cite{henaff2020data} where the instances in a batch are correlated, preventing the network from making meaningful predictions during inference, when cheating is not possible.
\item[Minimum required batch size:] Batch normalization does not work well with small batch size because the sample statistics computed from the batch fluctuate too wildly around the total population statistics which they are supposed to approximate. It would be nice to remove this limitation.
\item[Weird noise:] Batch normalization introduces noise in the training process because the mean and standard deviation of a given activation depend on the concrete choice of training samples in the batch, which changes each time. The noise might be seen as beneficial because of the regularizing effect of helping the model to get out of narrow loss minima, but of course the same effect could be achieved by just adding noise explicitly. The noise introduced by batch normalization thus has as its only benefit that it costs no extra computations, but I would argue that it is generally preferable to have noise that can be switched off, whose magnitude is independent of batch size and whose concrete shape is uncorrelated with the choice of instances in the batch.
\end{description}
Due to these issues, there are several attempts to partially fix them, or to get by without batch normalization. For example, Layer normalization\cite{ba2016layer} and Instance normalization\cite{ulyanov2016instance} normalize activation statistics along one or more tensor axes different from the batch axis. While this allows accumulation of batch statistics from multiple activations even at batch size 1, it has also led to problems; for example in \cite{karras2020analyzing} it was shown that the original StyleGAN exhibited blob-like artifacts because it ``cheats'' the normalization layer by creating a strong local spike in the activations that allows it to skew the statistics elsewhere. The proposed fix for the problem involved using not the actual activation statistics of the current instance, but rather the {\emph expected} statistics, and in that regard is conceptually similar to what this paper presents.
Batch renormalization \cite{ioffe2017batch} tries to enable smaller batch sizes by taking not just the mean and standard deviation of the current batch, but rather by maintaining a running average of these quantities over past batches. In fact, thinking about batch renormalization made me realize there is a better way to do what it is trying to do. It struck me as odd that a running average of standard deviations should be the right way to accumulate the statistics across several batches, considering that standard deviations aren't additive.

\section{Batchless normalization}
The gist of my solution is:
\begin{itemize}
\item Learn the mean $\my$ and standard deviation $\sigma$ of the activation $a$ by maximizing the likelihood that $a$ was sampled from a Gaussian distribution with the mean $\my$ and standard deviation $\sigma$.
\item To do this, include the negative log likelihood in the overall loss function.
\item Then, use the learned means and standard deviations to normalize the presumed distribution of $a$.
\item Then, as usual, scale and shift the activation to conform to a different, independently learned mean $\beta$ and standard deviation $\gamma$.
\end{itemize}
This can be done independently for each input to each normalization layer, or the statistics can be shared between different inputs, so that in a image processing convolutional network for example, all inputs belonging to the same channel but different pixels share their $\my$ and $\sigma$.

It is well known that a Gaussian distribution with the same mean and variance as given distribution, from which a large number of $a$-samples have been drawn independently, 
essentially maximizes the likelihood for all the samples to have been drawn from it among all Gaussian distributions. Hence, solving the optimization problem of finding the mean and variance of a normal distribution which maximizes the likelihood for samples of $a$ will yield the mean and variance of $a$, which can then be used to reparameterize $a$ to zero mean and unit variance.
We can maximize this likelihood by minimizing its negative logarithm. This makes the optimization problem compatible with gradient descent based optimizers because using the logarithm lets us write the formula as a sum of the contributions from individual $a$-samples, and making it negative turns the maximization objective into a minimization objective, as is customary for neural network optimization.

The likelihood for an activation $a_{\text{in}}$ to be drawn from a Gaussian distribution with mean $\my$ and standard deviation $\abs{\sigma}$ is
\begin{alignat}{2}
\mathcal{L} \left (\my, \abs \sigma ~\mline~ a_{\text{in}} \right )&&=&\frac1{\abs{\sigma}\sqrt{2 \pi}} \cdot e^{- \frac12 \left(\frac{a_{\text{in}} - \my}{\sigma}\right )^2}.\\
\intertext{The negative logarithm of that gives us as contribution to the loss:}
\textit{loss} &&\xleftarrow{+} & \lambda \cdot \left( \frac12 \left(\frac{a_{\text{in}} - \my}{\sigma} \right )^2 + \log \abs{\sigma}
 + \frac12 \log (2\pi)
 \right) 
\label{nosg-loss} \\
\intertext{And the activation is normalized as follows:}
a_{\textit{out}} && \leftarrow & \frac{a_{\text{in}} - \my}{\sigma} \cdot \gamma + \beta\label{nosg-acti}
\end{alignat}
Here, $\lambda$ is a learn rate multiplier that can be used to cause certain optimizers to learn $\my$ and $\sigma$ at a slower rate than other parameters\footnote{This does not work with all optimizers because some adapt to rescaled gradients by undoing the scaling}, which means that the history of activations influencing these parameters is longer, making them more steady and more similar to the parameters for the actual statistics for all training data. This is reasonable because internal covariance shift can be expected to occur slowly, and if the learned parameters should still deviate significantly from those of the actual distribution of the activation, they will generally do so consistently for large enough batch sizes, which means that momentum-based optimizers will make them converge quickly to near the actual mean and standard deviation despite the lower learning rate.

These formulas cannot be used as-is for backpropagation. On the one hand, the formula \vref{nosg-acti} should have no effect on the gradients with respect to $\my$ and $\sigma$, because formula \vref{nosg-loss} already does the job of learning these parameters, and that should not be messed up. On the other hand, formula \vref{nosg-loss} should have no effect on the gradients with respect to $a_{\textit{in}}$, because otherwise the preceding layer is incentivized to output activations that are close to $\my$, which in turn will cause $\sigma$ to shrink. This would presumably lead to a runaway effect where the activations converge towards $0$ over the course of training, until numerical precision problems arise because the numbers become denormalized (in the floating point sense, not in the statistical sense). 

To prevent these gradients from occurring in the wrong places, the stop-gradient function $\sg$ must be inserted into the formulas. During the forward pass 
$\sg$ behaves like the identity function, but its derivative is taken to be the $0$ function during backpropagation. This causes the argument of $\sg$ to be treated like a constant by the optimizer.

The formulas then become:
\begin{alignat}{2}
\textit{loss} &&\xleftarrow{+} & \lambda \cdot \left( \frac12 \left(\frac{\sg a_{\text{in}} - \my}{\sigma} \right )^2 + \log \abs{\sigma}
 + \frac12 \log (2\pi)
\right) 
\label{sg-loss} \\
a_{\textit{out}} && \leftarrow & \frac{a_{\text{in}} - \sg \my}{\sg \sigma} \cdot \gamma + \beta\label{sg-acti}
\end{alignat}

In order to utilize slightly cheaper instructions, it might be advantageous to not use $\sigma$ directly as a learnable parameter, but rather learn either $\log \sigma$ or $\sigma^{-1}$. This choice of how to parameterize $\sigma$ may interact with how the optimizer updates it. 

Depending on how $\sigma$ is parameterized by learnable parameters, the optimizer might set it to a negative value. Hence, its absolute value is used in the argument of the logarithm. 

The loss, as computed above for each individual activation, can simply added to the overall optimization objective. But since it tells us nothing about how well the network performs on the task it is being trained for, it should be kept separate for monitoring purposes and only added to the total loss in the end, like a regularization loss. Also, to keep the numbers at a reasonable order of magnitude, instead of summing the loss contributions from individual activations, it might be advantageous to take their per-layer average instead.

The parameters $\my$ and sigma $\sigma$, as well as $\gamma$ and $\beta$, have so far been talked about as if they affected only a single activation. But when not dealing with fully connected networks, it might make sense to share them across multiple activations, so that for example there is one set of them for each channel, but shared across the different grid positions in a convolutional network. This allows batchless normalization not only to replace batch normalization, but also techniques like instance normalization that normalize activation along other axes.

\subsubsection{Gauge}
The constant term $\frac12 \log (2\pi)$ can usually be omitted because it does not influence gradient computation. Only if you are interested in the actual negative log likelihood, you should keep it. But to have a more readable metric that tells us how well the learned distribution matches the actual distribution, we should use a different gauge, namely one so that the expected loss becomes zero if the distributions match.

Hence we need to subtract $g \eqdef \sg E_{x \sim \mathcal{N}(\my, \sigma^2)} 
\left(
	\lambda \cdot 
	\left( 
		\frac12 
		\left(
			\frac {x - \my}{\sigma}
		\right )
		^2 
		+ \log \abs{\sigma} 
		+ \frac12 \log (2\pi)
	\right)
\right)$ from the loss as computed in formula \vref{sg-loss}.

Since the $E$ is linear, we have
\begin{alignat}{2}
g &&= & \sg \left(\lambda \cdot \left( \frac12 \frac {E_{x \sim \mathcal{N}(\my, \sigma^2)} \left(x - \my\right)^2}{\sigma^2} + \log \abs{\sigma} + \frac12 \log (2\pi)\right)\right)
\label{sg-loss-gauge-1} \\
\intertext {Because $E_{x \sim \mathcal{N}(\my, \sigma^2)} \left(x - \my\right)^2$ is just the variance $\sigma^2$, this simplifies to}
g &&= & \sg 
\left(
	\lambda \cdot 
	\left( 
		\frac12 
		+ \log \abs{\sigma}
	    + \frac12 \log (2\pi)
	\right)
\right)
\label{sg-loss-gauge-2} 
\end{alignat}
Subtracting this from the loss contribution in formula \vref{sg-loss} gives us the correctly gauged loss contribution
\begin{alignat}{2}
\textit{loss} &&\xleftarrow{+} & \lambda \cdot 
\left( 
	\frac12 
	\left(
		\frac{\sg a_{\text{in}} - \my}{\sigma} 
	\right )
	^2 
	+ \log \abs{\sigma} 
	- \sg \log \abs{\sigma} 
 	- \frac12 
\right) 
\label{sg-loss-gauged} 
\end{alignat}
When aggregating these losses to obtain a single metric for how well the batchless normalization layer has learned the statistics parameters, we should prevent cancellation of positive and negative values from different activations. So, for the purpose of computing this metric we can take the square of the absolute value of these losses from individual activations before summing or averaging them. 

This metric may then be used to
\begin{itemize}
\item estimate the speed of internal covariance shift as the changes in statistics of the previous layer's outputs causes the learned statistics to lag behind the actual statistics.
\item control the batchless normalization layer's learning rate adaptively, as it can be made smaller when the statistics have been learned well and have stabilized, causing the parameters to be less dependent on the noisy selection of the last few batches.
\item assess how strongly data that have not been used during training (i.e. new data, or validation data, or data seen during inference) deviate from the training data.
\item monitor how fast the batchless normalization layers adapt to the actual statistics at the beginning of the training process. Although this should not be necessary because the normal way to go about this should be to initialize the statistics from a sample of the data.
\end{itemize}

\subsection{How this addresses the shortcomings of batch normalization}
All of the problems mentioned \vpageref{bnsc} are fixed:
\begin{description}
\item[Memory consumption:] It is now again possible to process the instances in a batch one by one while accumulating the gradients with respect to the parameters. When distributing the batch over multiple devices, no communication between them is necessary in between layers, only of course at the end of the batch for the parameter update.
\item[Implementation issues:] No more special treatment of the statistics parameters. They are learned using backpropagation, gradient descent and an optimizer like all other parameters.
\item[Train-test-discrepancy:] The network can and should be used just as it was trained. If you are concerned that the learned $\my$ and $\sigma$ are biased with respect to the statistics for the whole training set due to the last few batches, you can decrease $\lambda$ for the last few epochs. When the training is nearing convergence, there should be almost no covariance shift occuring, so the number of batches which significantly influence $\my$ and $\sigma$ can safely be increased.
\item[Cheating:] The instances in the batch are once more processed independently from each other, so cheating is not possible.
\item[Minimum required batch size:] It works even with a batch size of $1$.
\item[Weird noise:] It appears that the network output fluctuates significantly less than with batch normalization or without any normalization, see the experiment \vpageref{fluctu}. Any noise that remains is uncorrelated to the current batch because $\my$ and $\sigma$ do not depend on it.
\end{description}

\subsection{initialization}
If the actual statistics of the batchless normalization layer inputs differ too much from those it has learned or been initialized with, it will be slow or even unable to adapt. For this reason, it is recommended to initialize the $\my$ and $\sigma$ parameters by running the untrained network on a relatively small sample of the training data, observe the statistics, and initialize the batchless normalization layers accordingly. Since initializing the first batchless normalization layer changes the statistics of its output, and therefore the statistics of the input for the next batchless normalization layer, there need to be multiple passes over the sample of training data. The sample should be large enough to be representative of the distribution of lowest-level features of the data, but it need not be representative of highest level features because the network has not learned those yet as its linear layers are randomly initialized.

\subsection{Possible problems with batchless normalization}
There are some slight disadvantages compared to batch normalization:
\begin{itemize}
\item The statistics parameters are always lagging behind by one batch, whereas in batch normalization they are always up-to date (but noisy because they depend only on that batch) and in batch renormalization they are at least influenced by the current batch's statistics. But based on the reasonable assumption that the distributions shift only slightly between one batch and the next, and further assuming the learned $\my$ and $\sigma$ parameters approximate the population statistics already quite well, this seems like a small price to pay, and also a necessary one to prevent the problems that come from lack of independent processing of the instances inside a batch.
\item In case that the actual statistics -- after initialization of the network weights, but before any training -- differ significantly from the initial values of the $\my$ and $\sigma$  parameters, several iterations will be spent on adapting the parameters to the actual statistics before the network can start learning something useful. However, it is already standard practice for unnormalized networks and several alternatives to batch normalization to initialize the weights to carefully chosen values. 
Also, with batchless normalization we have the choice whether we want to carefully initialize the linear layer weights, or rather initialize the $\my$ and $\sigma$ parameters so that they conform to the statistics resulting from the arbitrarily initialized linear layer weights, as described above.
\end{itemize}

\subsection{Migrating to batchless normalization}
If you want to use batchless normalization for future training on a model that is already pretrained, it is simple to convert it:
If the model already uses batch normalization or batch renormalization, its parameters will contain the means and standard deviations that were determined when it was trained. Simply use these to initialize the $\my$ and $\sigma$  parameters of batchless normalization (along with possibly the $\beta$ and $\gamma$  parameters, if batch normalization is followed by denormalization).

If the model did not previously use any form of batch normalization, run it on a large sample of training data to determine the mean and standard deviation of each activation you want to normalize. Then, use these to initialize $\my$ and $\sigma$ for each activation, and initialize $\beta$ and $\gamma$ to be equal to $\my$ and $\sigma$, respectively, in order to ensure that the network still computes the same function. Unlike with initialization, it is not necessary here to do multiple passes, but it is more important that high-level features are well represented in the sample because the model has presumably already learned these.

\subsection{Implementation}
An implementation of batchless normalization for TensorFlow is available under
\url{https://github.com/ichteltelch/Batchless}.
The repository also contains the code for running the experiment in \vref{cifar10} as an usage example.

\section{Experiments}
Here, I present some small experiments to demonstrate that batchless normalization works, is competitive with batch (re)normalization, and makes the network outputs during training more steady. % and quickly learns the statistics parameters even in the face of bad initialization.

\subsection{Convergence and Stability}
\label{fluctu}
This first experiment is meant to demonstrate that that batchless normalization works as expected in that it does not break the network and instead speeds up training comparable to batch (re)normalization. I also investigate a finding that I did not anticipate, 
namely that after convergence the network output for a given input fluctuates less from batch to batch when batchless normalization is used. I take this as an indication that batchless normalization is less dependent on the choice of instances for each batch, which I should have expected, and it is less prone to jump out of a local loss minimum that it finds, which may be considered a negative thing, but as said before can always be remedied by explicitly injecting noise and thus allows more control over how much noise is used.

The structure of the neural network used in this experiment can be found in \vref{fluctuTop}

\subsubsection{Data set}
\begin{figure}[htp]
\center\includegraphics[width=0.5\textwidth]{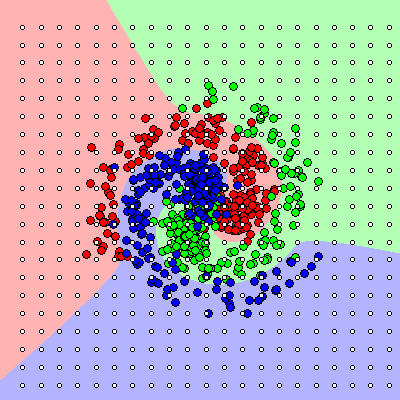}
\caption{Structure of the example problem. Not all data points are shown. Background color indicates the association from coordinates to class label as learned by one neural network instance.}
\label{dataset1}
\end{figure}

The data set I use for this experiment is a classification task where the input consists of x and y coordinates of a point and the output
is one of three classes. For each class I use 20~000 training data points and 4~000 validation data points. The data points are arranged in three intertwined, randomly diffused spirals, one for each class; see Figure \vref{dataset1}.

The colorless circles mark the fluctuation measurement sites which are inputs for which I measure the fluctuation of the outputs, in order to determine how stable the convergence is.
\subsubsection{Training and evaluation strategy}
The neural networks are trained with batches of various sizes in order to observe the effect of batch size. 
The learning rate is 0.01 for all batch sizes, to enable better comparability of the results. 
The AMSGrad optimizer\cite{reddi2019convergence} is used.
Each $n\cross m$ weight matrix is initialized from a uniform distribution of mean $0$ and width $\sqrt{\frac{2}{n+m}}$.
For the batchless normalization layers, $\lambda$ is $0.1$.
The batches are randomly drawn subsets of the training data points.
When the median training loss of the last 15 batches has not reached a new low for 1~000 batches, the network is assumed to have converged.
For another 1~000 batches, the network's outputs (which are probability distributions over the three classes) are then recorded on a regular grid in the input space.
Then the fluctuation is computed as the average relative entropy from each of these outputs to the mean output\footnote{Using the mean as the center from which to measure the deviations is justified because the pointwise mean of several probability distributions minimizes the summed relative entropies to all of them.} for each fluctuation measurement site.
Validation loss is computed after the 1~000 batches to measure fluctuation have been recorded while continuing to train the network. For batch normalization and batch renormalization, the statistics parameters used for normalization are determined using the entire training data set before validation loss is computed. The loss reported here is that computed by the final loss layer and does not include the contributions from the batchless normalization layers.
\subsubsection{Results}
Table \vref{valloss} shows the validation losses. For all normalization strategies, the loss tends to decrease with increasing batch size. For all batch sizes, all variants of batchless normalization outperform batch normalization and no normalization, but batch renormalization is able to catch up at larger batch sizes. Batchless normalization appears to work best when the standard deviations are parameterized via their logarithm (BlNlog).

Very small batch sizes are harmful for batch (re)normalization, making the loss larger than it would be without the attempt to normalize. This is to be expected because the batch statistics are a bad approximation to the population statistics for small sample sizes. Batchless normalization, in contrast, is an improvement over no normalization even for batch size $1$.\footnote{This may not really be relevant, because using larger batch size is usually advantageous and batchless normalization enables free choice of batch size regardless of memory constraints.}

The average losses computed over all training data, which are not shown here, are very similar to the validation losses; in fact for some reason they are always around one percent higher than the corresponding validation loss. This probably means that I used too many training examples considering the complexity of the problem and of the neural networks used here. The generalization ability is thus not tested by this experiment.
\begin{table}[htp]
\center
\begin{tabular}{r|llllll}
Batch size   & 	Ø           & 	BN          & 	BRN        & 	BlN          & 	BlNlog       & 	BlNinv\\
\hline
1 & 	$0.9270719$ & 	-           & 	-          & 	$0.43877026$ & 	$0.40771368$ & 	$0.40555158$\\
2 & 	$0.7150986$ & 	$1.0413455$ & 	$0.9223012$ & 	$0.36359715$ & 	$0.3537329$ & 	$0.3559611$\\
4 & 	$0.54393834$ & 	$0.85783184$ & 	$0.40026915$ & 	$0.32604578$ & 	$0.32921416$ & 	$0.32076585$\\
8 & 	$0.4349341$ & 	$0.53910136$ & 	$0.31462723$ & 	$0.30360582$ & 	$0.3036276$ & 	$0.29853466$\\
16 & 	$0.3791601$ & 	$0.4020436$ & 	$0.28705415$ & 	$0.2898234$ & 	$0.286525$ & 	$0.28917935$\\
32 & 	$0.32971925$ & 	$0.32011843$ & 	$0.27646378$ & 	$0.27794752$ & 	$0.2749193$ & 	$0.27682093$\\
64 & 	$0.3014693$ & 	$0.28662997$ & 	$0.26976454$ & 	$0.27084148$ & 	$0.2703774$ & 	$0.27034158$\\
\end{tabular}
\caption{Validation loss for different batch sizes and different types of normalization, averaged over 100 test runs}\label{valloss}
\end{table}

Table \vref{tillconv} shows the number of batches it took to converge (in the sense mentioned above). 
Note that faster convergence is only better if the loss reached in the end is comparable, otherwise it just means the algorithm gave up earlier because the loss could not be decreased further. With this in mind, we see that the convergence of batchless normalization takes only slightly longer while mostly also yielding a slightly better loss than for example batch renormalization. Ordinary batch normalization converges fastest, but loss-wise it is only an improvement on having no normalization, and that only for larger batch sizes.
\begin{table}[thp]
\center
\begin{tabular}{r|rrrrrr}
Batch size & Ø & 	BN     & 	BRN    & 	BlN      & 	BlNlog       & 	BlNinv\\
\hline
1   & 	$1962$ & 	-      & 	-      & 	$3135$ & 	$3369$ & 	$3666$\\
2 & 	$2485$ & 	$1907$ & 	$2696$ & 	$3498$ & 	$3582$ & 	$3456$\\
4 & 	$2679$ & 	$1754$ & 	$3516$ & 	$3325$ & 	$3151$ & 	$3236$\\
8 & 	$2933$ & 	$2605$ & 	$3254$ & 	$3114$ & 	$3080$ & 	$2815$\\
16 & 	$3088$ & 	$2549$ & 	$2758$ & 	$2870$ & 	$2774$ & 	$2910$\\
32 & 	$2893$ & 	$2617$ & 	$2806$ & 	$2942$ & 	$2891$ & 	$2944$\\
64 & 	$3204$ & 	$2743$ & 	$2910$ & 	$2954$ & 	$3076$ & 	$2880$\\
\end{tabular}
\caption{Batches until convergence for different batch sizes and different types of normalization, averaged over 100 test runs and rounded}\label{tillconv}
\end{table}

Table \vref{tabfluctu} shows the fluctuation of the network output when training continues after convergence. For all normalization strategies, the fluctuation tends to decrease with increasing batch size, which is to be expected because the gradients become less noisy. For all batch sizes, all of the batchless normalization variants are clearly more stable than the other normalization strategies.
\begin{table}[htp]
\center
\begin{tabular}{r|llllll}
Batch size & 	Ø    & 	BN          & 	BRN        & 	BlN          & 	BlNlog       & 	BlNinv\\
\hline
1 & 	$0.17402601$ & 	-          & 	-          & 	$0.09130978$ & 	$0.076311015$ & $0.074348584$\\
2 & 	$0.19333616$ & 	$0.11037045$ & 	$0.09140002$ & 	$0.06586651$ & 	$0.05923214$ & 	$0.058804046$\\
4 & 	$0.1864283$ & 	$0.06686098$ & 	$0.082503244$ & $0.04783394$ & 	$0.044846766$ & $0.045572363$\\
8 & 	$0.14548533$ & 	$0.05250681$ & 	$0.05716909$ & 	$0.035242695$ & $0.03462934$ & 	$0.035029452$\\
16 & 	$0.103627704$ & $0.044835586$ & $0.040057465$ & $0.025415251$ & $0.024552444$ & $0.025088605$\\
32 & 	$0.07482936$ & 	$0.035840534$ & $0.026805528$ & $0.01831559$ & 	$0.018427724$ & $0.018823238$\\
64 & 	$0.05339188$ & 	$0.028317539$ & $0.019306945$ & $0.013491537$ & $0.013304713$ & $0.013664371$\\
\end{tabular}
\caption{Average output fluctuation for different batch sizes and different types of normalization, averaged over 100 test runs}\label{tabfluctu}
\end{table}
It might have been informative to look at not only the fluctuation of the outputs, but also at the variance of the weights or of the gradients of loss with respect to the weights during training. I conjecture we would have seen something similar. The differences in stability could be used to select the highest stable leaning rate for each scenario. This would allow us to compare the actual amount of computation needed in a practical setting. But for this experiment, I used a uniform learning rate, which makes especially the numbers in table 
\vref{tabfluctu} more comparable among each other.

\skipall{
\subsubsection{Further experiments}
The network used in the experiment was very shallow, but the benefits of batch normalization are generally more pronounced in deeper networks.
I did some more experiments with a deeper network (around 20 layers), and also with adaptive learning rate control. It seemed that in this setting, ordinary batch (re)normalization at larger batch sizes was better than batchless normalization, although the latter still vastly outperformed no normalization, and thus remains an attractive alternative in situations where the aforementioned shortcomings of batch (re)normalization are prohibitive. However, I do not include the numbers here because I do not really believe these results to be more than a hint, seeing as the toy problem that the networks were trained to solve is far too easy, and using deeper networks did not really reduce the loss as compared to the shallow ones in the first experiment.

Beyond the perhaps oversimplistic toy problem, I have successfully employed batchless normalization in my day job, but I don't think the bosses would like it if I gave details about that.
}

\subsection{CIFAR-10 classification}
\label{cifar10}
In this experiment, the CIFAR-10 dataset is used to compare various flavors of batchless normalization to batch normalization and to no normalization at all. The structure of the neural networks used is described in detail in \vref{cifarTop}. For the initialization of the BlN layers, the first $1000$ training samples were used. The Adam optimizer was used to train the networks.

The models were trained for 60 epochs with different batch sizes. The maximum validation accuracies over all epochs are reported in table \vref{cifarAccu}. The minimum validation losses are likewise reported in table \vref{cifarLoss}.

\begin{table}[htp]
\center
\begin{tabular}{r|llllll}
Batch size & 	Ø    & 	BN           & 	BlN          & 	BlNlog       & 	BlNinv\\
\hline
1 & 	$0.3860$ & -        & $0.7777$ & $0.7747$ & $0.7438$ \\
2 & 	$0.5430$ & $0.2680$ & $0.7794$ & $0.7826$ & $0.7695$ \\
4 & 	$0.6478$ & $0.7766$ & $0.7775$ & $0.7826$ & $0.7800$ \\
8 & 	$0.7127$ & $0.7976$ & $0.7855$ & $0.7789$ & $0.7824$ \\
16 & 	$0.7394$ & $0.7959$ & $0.7846$ & $0.7812$ & $0.7832$ \\
32 & 	$0.7622$ & $0.7900$ & $0.7849$ & $0.7808$ & $0.7871$ \\
64 & 	$0.7681$ & $0.7844$ & $0.7870$ & $0.7734$ & $0.7782$ \\
128 & 	$0.7688$ & $0.7716$ & $0.7767$ & $0.7871$ & $0.7780$ \\
256 & 	$0.7664$ & $0.7790$ & $0.7732$ & $0.7679$ & $0.7767$ \\
512 & 	$0.7596$ & $0.7737$ & $0.7723$ & $0.7734$ & $0.7666$ \\
1024 & 	$0.7627$ & $0.7653$ & $0.7550$ & $0.7573$ & $0.7601$ \\
\end{tabular}
\caption{Validation accuracy for different batch sizes and different types of normalization, maximum over 60 epochs}\label{cifarAccu}
\end{table}

\begin{table}[htp]
\center
\begin{tabular}{r|llllll}
Batch size & 	Ø    & 	BN           & 	BlN          & 	BlNlog       & 	BlNinv\\
\hline
1 & 	$1.9942$ &  -       & $0.7137$ & $0.7488$ & $0.7914$ \\
2 & 	$1.5579$ & $2.0516$ & $0.7436$ & $0.6943$ & $0.7464$ \\
4 & 	$1.4061$ & $0.6727$ & $0.7451$ & $0.7184$ & $0.7212$ \\
8 & 	$1.1322$ & $0.6028$ & $0.7102$ & $0.7495$ & $0.7353$ \\
16 & 	$1.0385$ & $0.6078$ & $0.7054$ & $0.7164$ & $0.7109$ \\
32 & 	$0.9370$ & $0.6307$ & $0.7260$ & $0.7416$ & $0.6922$ \\
64 & 	$0.8542$ & $0.6674$ & $0.7112$ & $0.7222$ & $0.7243$ \\
128 & 	$0.8678$ & $0.7208$ & $0.7258$ & $0.6892$ & $0.7219$ \\
256 & 	$0.8120$ & $0.7068$ & $0.7307$ & $0.7214$ & $0.6985$ \\
512 & 	$0.7717$ & $0.7195$ & $0.6977$ & $0.6904$ & $0.7412$ \\
1024 & 	$0.7355$ & $0.7331$ & $0.7329$ & $0.7154$ & $0.7192$ \\
\end{tabular}
\caption{Validation loss for different batch sizes and different types of normalization, minimum over 60 epochs}\label{cifarLoss}
\end{table}

The evolution of validation loss and accuracy over the 0 epochs, of which these tables only report the extremal values are plotted in \vref{cifar10Plots}.

As it can be seen, batchless normalization is mostly on par with batch normalization, except at very small batch sizes, where it alone is viable, ant at medium batch sizes around 8 to 128, where batch normalization performs slightly better on this data set. Among the variants of batchless normalization tried, the one that parametrizes the variance using its logarithm seems to be slightly more reliable.

\skipall{
\section{How fast are the statistics parameters learned?}
In this experiment, I investigate how fast batchless normalization adapts to the actual statistics of the activaitons if the initial statistics are very different from what the batchless normalization layer has learned. 

To use the simplest possible case, I use the trivial and exactly solvable task of learning the identity function on standard normal distributed scalars. The network has 10 hidden layers with two ReLu neurons each. Batchless normalization occurs between each linear layer and the following ReLu nonlinearity layer. The weights of each hidden linear layer\footnote{That is, all that are followed by a nonlinearity layer; the last linear layer that produces the output is initialized with reasonable weights ad zero bias.} are initialized to be on the order of $e^d$, where $d$ is sampled for each layer from the uniform distribution on $[-10; 10]$. The biases of each linear layer are sampled uniformly from $[-100; 100]$.
}

\section{Conclusion and Further work}
This paper merely presented an idea and a proof of concept. More testing using a wider range of tasks, architectures, model sizes, and hyperparameters should probably be done. 

The experimental results so far suggest that batchless normalization can hold its own against batch renormalization in terms of convergence speed and loss, is better than batch normalization at least for shallow networks, and beyond that enables normalization with batch sizes too small for both previous normalization strategies to be applicable.

The interaction between the optimizer used and various ways to parameterize the standard deviations should be mapped, as there seems to be some importance to it. For example, with the the AMSGrad optimizer I used here, the history of absolute gradient magnitudes is relevant, so it matters how the standard deviations are represented in the parameter vector.

It has been observed that weight decay and batch normalization interact in strange ways when it comes to training dynamics, see for example \cite{lobacheva2021periodic}. In brief, this is because weight decay causes the distribution of activations coming out of the previous layer to shrink towards zero, which has then no effect on the subsequent layers (and thus the loss) because it is undone by normalization. This collapse of the distribution is counteracted by the batch normalization's effect of making the gradient vector perpendicular to the weight vector so that parameter updates lengthen the weight vector, which leads to periodic destabilization. Batchless normalization is presumably also affected by this phenomenon, but exactly how it plays out here I did not investigate. \skipall{ Should it prove to be detrimental, it can presumably be counteracted after each training step by normalizing each row of the weight matrix of the preceding linear layer to be a unit vector, and then absorbing the scale factor used in this vector normalization into the $\my$ and $\sigma$ of the batchless normalization layer so that the function computed by the network does not change.}

The increased stability might allow us to choose a larger stable learning rate, leading to faster training. I did not investigate this possiblility.

Instead of learning the $\my$ and $\sigma$ parameters more or less directly, it would also be possible to compute them using a small auxiliary neural network from e.g. positional encodings. For certain types of spatial data, it may be useful to be able to handle activation statistics that depend on location. This is just an idea about what might be possible once the statistics parameters no longer require special treatment, and I have not researched it any further.

\section{Impact}
While it remains to be seen what the effect of the reduced noise is on the training dynamics of large models, even if batchless normalization should turn out to perform quite similar to other normalization schemes, one possibility is opened up for certain:
Batchless normalization will enable more people and institutions to train larger models (utilizing normalization across instances) on their own hardware while allowing free choice of batch size. 
It will of course take longer than if they had better hardware, as the amount of compute is not significantly changed, but at least it will be possible where before it wasn't. 
Thus batchless normalization will hopefully contribute to the democratization of AI research. 
Then again, it might also move the maximum model size/efficiency Pareto frontier for models trained using normalization by the top few most resource-rich entities and hence have the opposite effect at the level of very large models.
However, alternatives to batch normalization that normalize only within a single instance, such as layer normalization, are already well established for models so large that batch normalization is prohibitive. As these alternatives seem to work reasonably well, it is questionable seen whether cross-instance normalization will even be adopted for these use case now that it is feasible.

\section*{References}
\bibliographystyle{unsrt}
\bibliography{batchless}
\appendix
\section{Network Structure for Experiment ``Convergence and Stability''}
\label{fluctuTop}
The network used is a feed forward network with the following structure:
\begin{itemize}
\item A dense linear layer with 50 outputs
\item Possibly a normalization layer (None (Ø), batch normalization (BN), batch renormalization (BRN), or batchless normalization with the standard deviations parameterized either directly (BlN) or by their logarithms (BlNlog) or their inverses (BlNinv)).
\item A nonlinearity layer, employing the ISRLU activation function \cite{carlile2017improving}
\item A dropout layer
\item Repeated twice:
\begin{itemize}
\item A dense linear layer with 40 outputs
\item Possibly another normalization layer like the first one
\item A nonlinearity layer, employing the ISRLU activation function
\item A dropout layer
\end{itemize}
\item A dense linear layer with 3 outputs
\item A SoftMax layer
\end{itemize}
The parameter for the ISRLU neurons is $4$. The probability for the dropout layers is $0.9$. The weight matrices for the dense linear layers 
are regularized with $L_2$ weight decay of strength $10^{-6}$.

\section{Network Structure for Experiment ``CIFAR-10 classification''}
\label{cifarTop}
The network used is a convolution network with the following structure:
\begin{itemize}
\item Possibly a normalization layer (None (Ø), batch normalization (BN), or batchless normalization with the standard deviations parameterized either directly (BlN) or by their logarithms (BlNlog) or their inverses (BlNinv)). The statistics parameters are shared across all inputs that belong to the same channel but different pixels.
\item A $7 \cross 7$ convolutional layer with same-paadding and 64 output channels
\item A leaky ReLU nonlinearity with a slope of $0.3$
\item Possibly another normalization layer like the first one
\item A dropout layer with probability $0.25$
\item A $2 \cross 2$ max-pooling payer
\item A $5 \cross 5$ convolutional layer with same-paadding and 64 output channels
\item A leaky ReLU nonlinearity with a slope of $0.3$
\item Possibly another normalization layer like the first one
\item A dropout layer with probability $0.25$
\item A $2 \cross 2$ max-pooling payer
\item A $3 \cross 3$ convolutional layer with same-paadding and 64 output channels
\item A leaky ReLU nonlinearity with a slope of $0.3$
\item Possibly another normalization layer like the first one
\item A dropout layer with probability $0.25$
\item A $2 \cross 2$ max-pooling payer
\item A flattening layer
\item A dense linear layer with 50 outputs
\item A leaky ReLU nonlinearity with a slope of $0.3$ 
\item Possibly another normalization layer like the first one, except that the statistics parameters are no longer shared because there are no more distinct pixels.
\item A dropout layer with probability $0.25$
\item A dense linear layer with 50 outputs
\item A leaky ReLU nonlinearity with a slope of $0.3$ 
\item Possibly another normalization layer
\item A dropout layer with probability $0.25$
\item A dense linear layer with 10 outputs
\item A softmax layer

\section{Plots of the training dynamics for Experiment ``CIFAR-10 classification''}
\label{cifar10Plots}
\begin{figure}[htp]
\center
\includegraphics[width=0.45\textwidth]{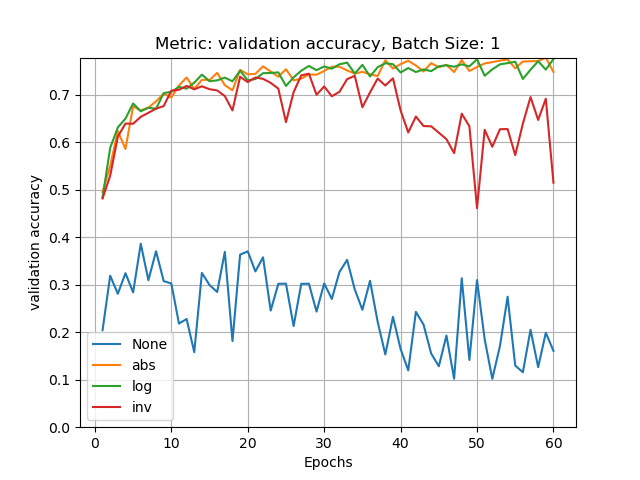}\includegraphics[width=0.45\textwidth]{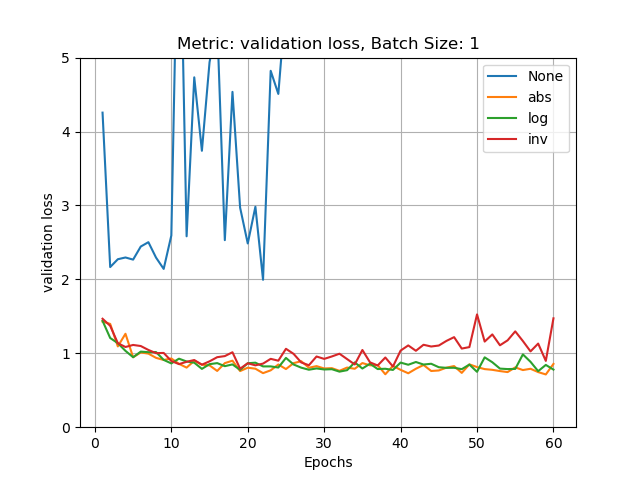}
\caption{For batch batch size 1. Note that batch normalization is not applicable here.}
\end{figure}

\begin{figure}[htp]
\center
\includegraphics[width=0.45\textwidth]{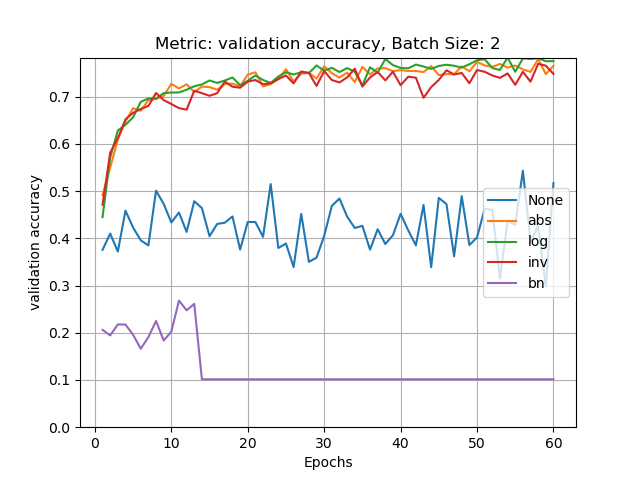}\includegraphics[width=0.45\textwidth]{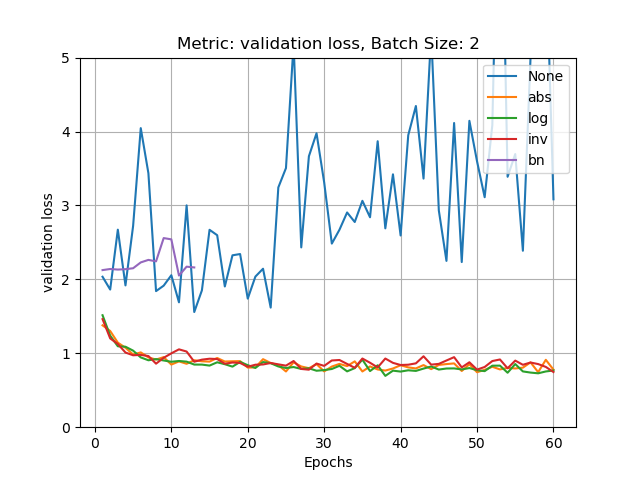}
\caption{For batch batch size 2. The graphs for batch normalization flatline or disappear, respectively, because training diverged to NaN values.}
\end{figure}

\begin{figure}[htp]
\center
\includegraphics[width=0.45\textwidth]{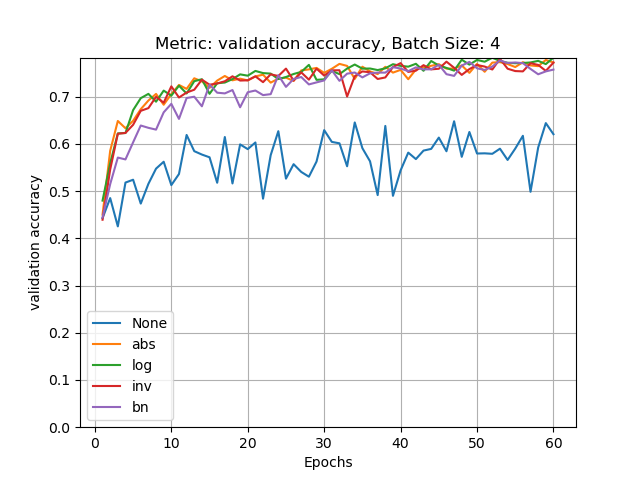}\includegraphics[width=0.45\textwidth]{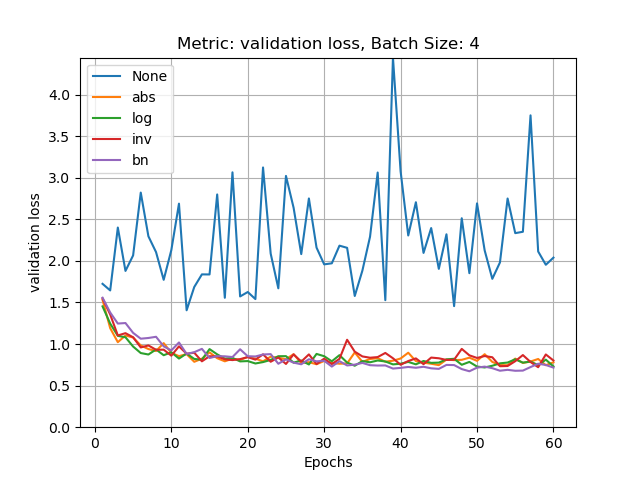}
\caption{For batch batch size 4. Initially, batchless normalization converges faster, but is later slightly overtaken by batch normalization.}
\end{figure}

\begin{figure}[htp]
\center
\includegraphics[width=0.45\textwidth]{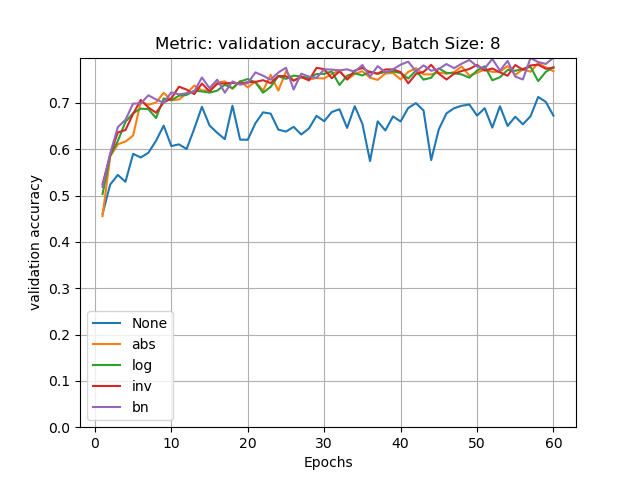}\includegraphics[width=0.45\textwidth]{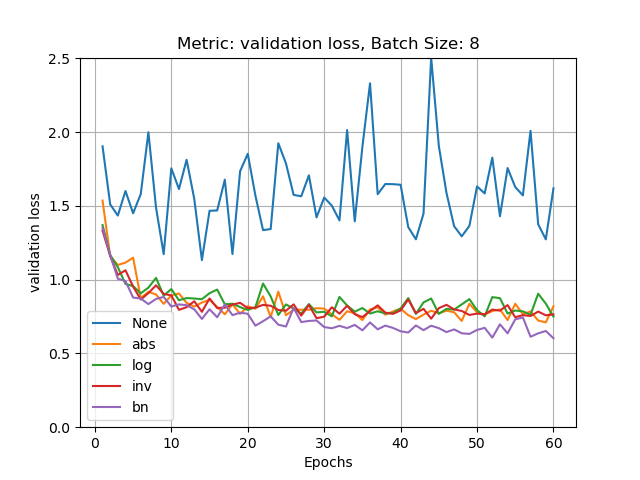}
\caption{For batch batch size 8. Batch normalization is better, but all networks using any kind of normalization learn vastly better than with no normalization at all.}
\end{figure}

\begin{figure}[htp]
\center
\includegraphics[width=0.45\textwidth]{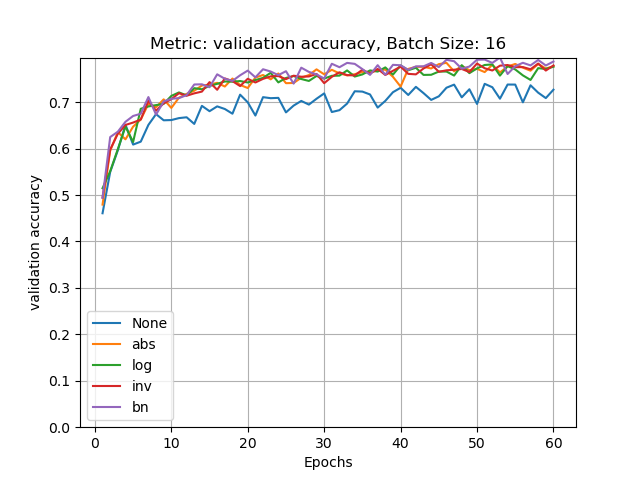}\includegraphics[width=0.45\textwidth]{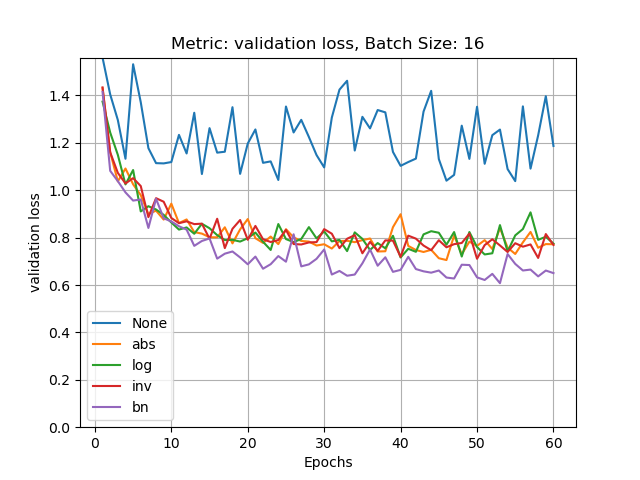}
\caption{For batch batch size 16.}
\end{figure}

\begin{figure}[htp]
\center
\includegraphics[width=0.45\textwidth]{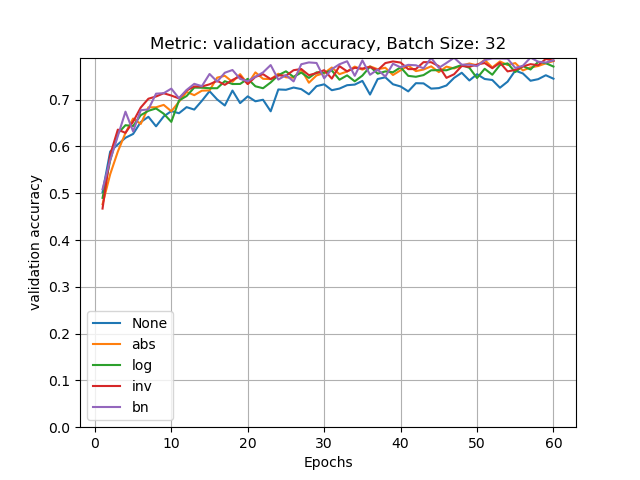}\includegraphics[width=0.45\textwidth]{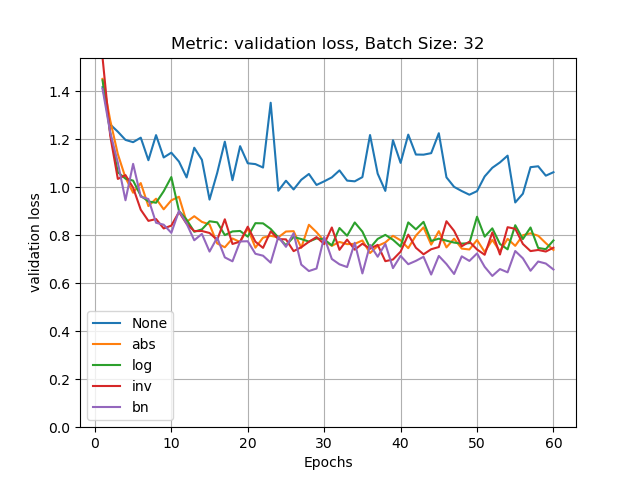}
\caption{For batch batch size 32.}
\end{figure}

\begin{figure}[htp]
\center
\includegraphics[width=0.45\textwidth]{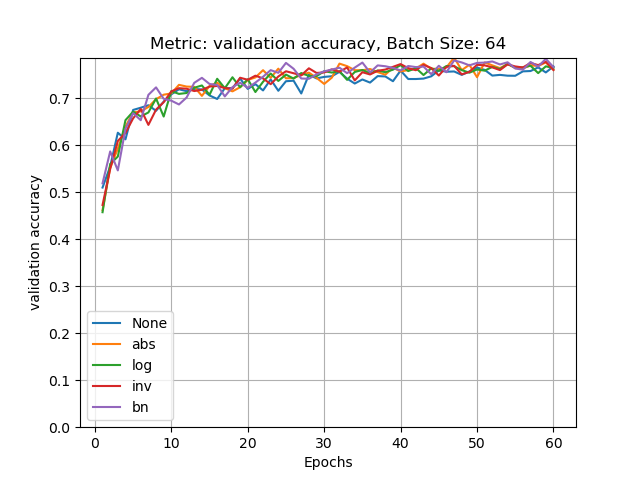}\includegraphics[width=0.45\textwidth]{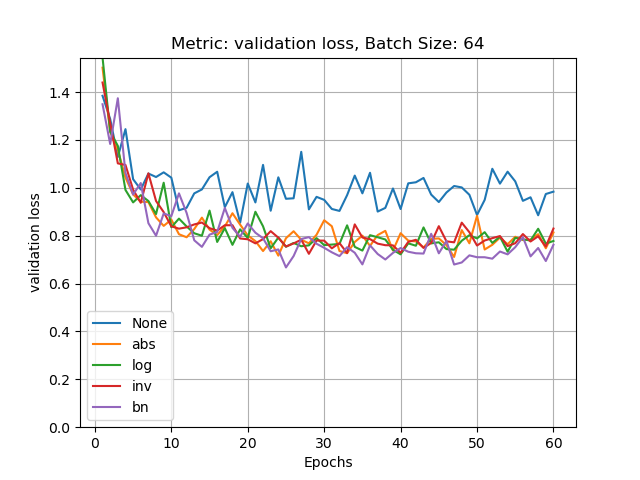}
\caption{For batch batch size 64.}
\end{figure}

\begin{figure}[htp]
\center
\includegraphics[width=0.45\textwidth]{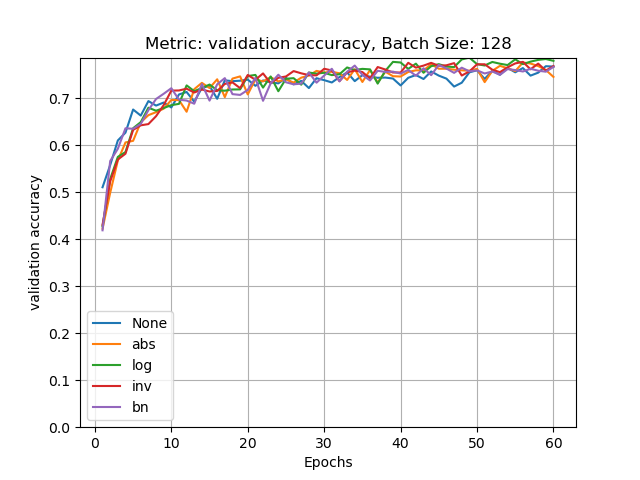}\includegraphics[width=0.45\textwidth]{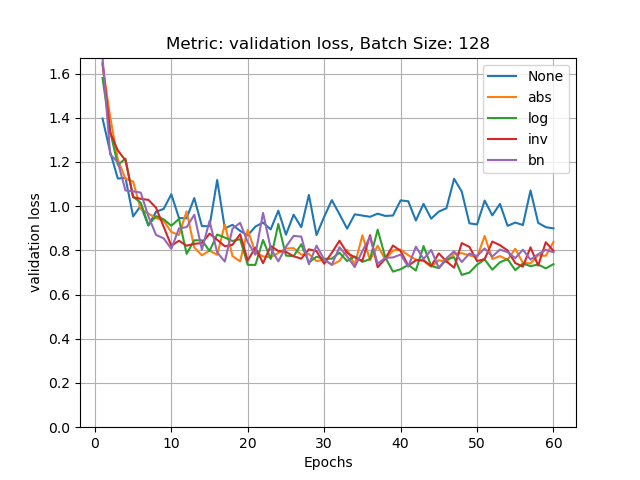}
\caption{For batch batch size 128.}
\end{figure}

\begin{figure}[htp]
\center
\includegraphics[width=0.45\textwidth]{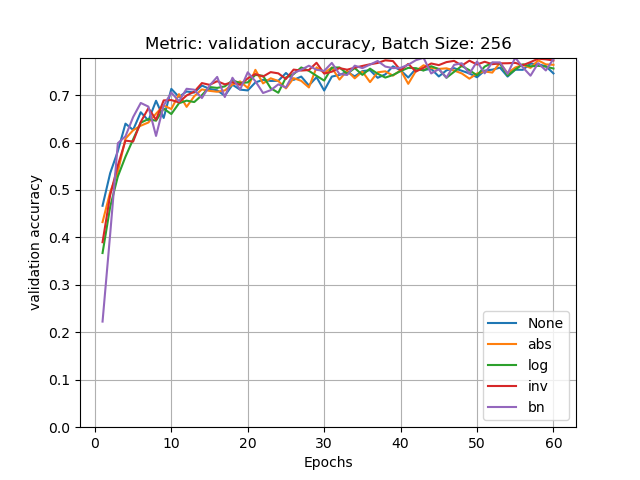}\includegraphics[width=0.45\textwidth]{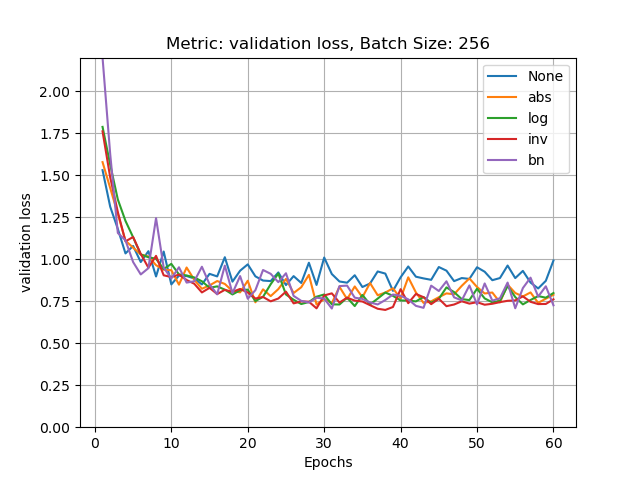}
\caption{For batch batch size 256.}
\end{figure}

\begin{figure}[htp]
\center
\includegraphics[width=0.45\textwidth]{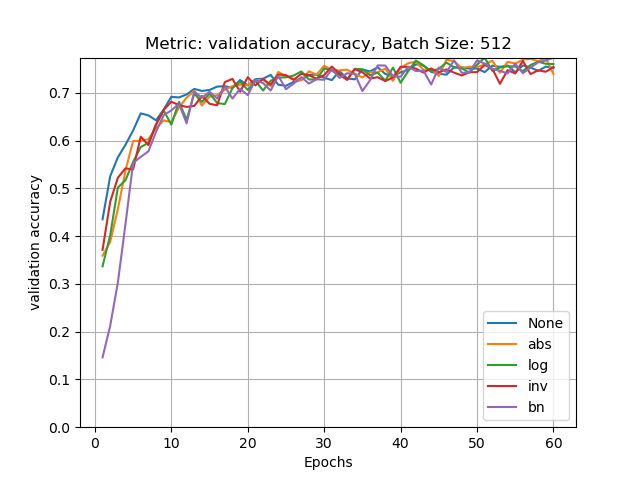}\includegraphics[width=0.45\textwidth]{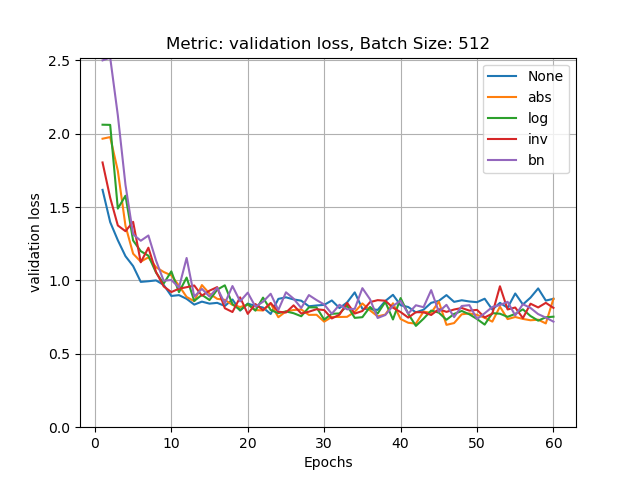}
\caption{For batch batch size 512.}
\end{figure}

\begin{figure}[htp]
\center
\includegraphics[width=0.45\textwidth]{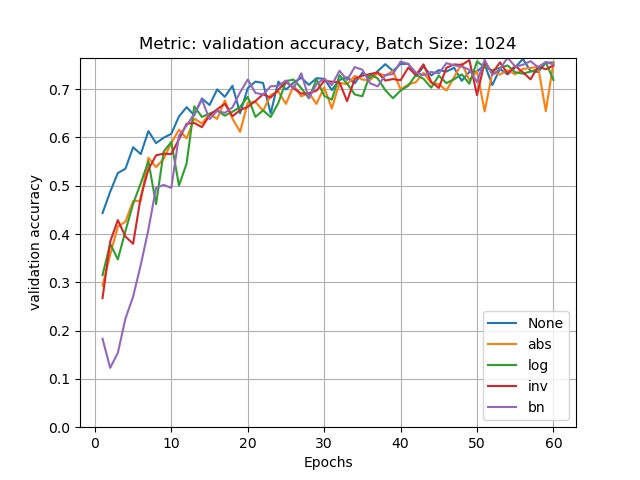}\includegraphics[width=0.45\textwidth]{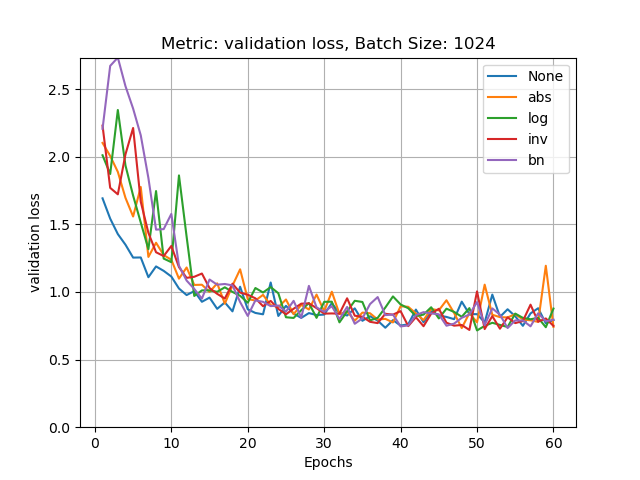}
\caption{For batch batch size 1024. Whether normalization is used or not seems to make little difference here; but for deeper networks, normalization can be expected to remain vital even at very high batch sizes. }
\end{figure}

\end{itemize}

\end{document}